\documentclass{article}

\PassOptionsToPackage{authoryear,round}{natbib}

    \usepackage[preprint]{neurips_2025}

\usepackage[utf8]{inputenc} 
\usepackage[T1]{fontenc}    
\usepackage{hyperref}       
\usepackage{url}            
\usepackage{booktabs}       
\usepackage{amsfonts}       
\usepackage{nicefrac}       
\usepackage{microtype}      
\usepackage{xcolor}         
\hypersetup{
    colorlinks=true,   
    citecolor=blue,    
    linkcolor=blue,    
    urlcolor=blue      
}

\usepackage{multirow}
\usepackage{pifont}
\usepackage{bbding}
\usepackage{array}
\usepackage{tikz}
\usepackage{colortbl}   
\usepackage{epigraph}
\usepackage{wrapfig}
\usepackage{makecell}

\usepackage[most]{tcolorbox}
\usepackage{caption}

\usepackage{soul}

\definecolor{code}{HTML}{fefadd}
\definecolor{result}{HTML}{ffe3ec}

\definecolor{code_text}{HTML}{dcd28e}
\definecolor{result_text}{HTML}{e0a7ba}
\definecolor{think_text}{HTML}{c9d6f1}

\usetikzlibrary{tikzmark}
\newcommand{\textboxStyle}[2]{%
    \tikzmarknode[draw=#1,thick,fill=white,rounded corners=2pt,inner sep=2pt]{boxnode}{#2}%
}

\newcommand{\thinkStart}{\textboxStyle{think_text}{\textbf{\textcolor{think_text}{<think>}}}}
\newcommand{\thinkEnd}{\textboxStyle{think_text}{\textbf{\textcolor{think_text}{</think>}}}}

\newcommand{\resultStart}{\textboxStyle{result_text}{\textbf{\textcolor{result_text}{<execution\_results>}}}}
\newcommand{\resultEnd}{\textboxStyle{result_text}{\textbf{\textcolor{result_text}{</execution\_results>}}}}

\newcommand{\codeStart}{\textboxStyle{code_text}{\textbf{\textcolor{code_text}{<code>}}}}
\newcommand{\codeEnd}{\textboxStyle{code_text}{\textbf{\textcolor{code_text}{</code>}}}}

\definecolor{bgcolor}{HTML}{F1F4F7}

\title{\Large{SciMaster: Towards General-Purpose Scientific AI Agents} \\
\large\textit{Part I. X-Master as Foundation — Can We Lead on Humanity’s Last Exam?}}

\author{%
  \textbf{Jingyi Chai\textsuperscript{1}\thanks{Equal Contributions. The ordering was randomized via a dice roll.}}\quad \textbf{Shuo Tang\textsuperscript{1}\footnotemark[1]} \quad \textbf{Rui Ye\textsuperscript{1}\footnotemark[1]}  \quad \textbf{Yuwen Du\textsuperscript{1}\footnotemark[1]} \quad \textbf{Xinyu Zhu\textsuperscript{1}} \quad 
  \textbf{Mengcheng Zhou\textsuperscript{1}} \\ \quad \textbf{Yanfeng Wang\textsuperscript{1}} \quad \textbf{Weinan E\textsuperscript{1,2}} \quad \textbf{Yuzhi Zhang\textsuperscript{2}} \quad \textbf{Linfeng Zhang\textsuperscript{2}} 
  \quad \textbf{Siheng Chen\textsuperscript{1}}\\
  \textsuperscript{1} School of Artificial Intelligence, Shanghai Jiao Tong University \quad
  \textsuperscript{2}  DP Technology\\
  \textcolor{blue}{X-Master:} \url{https://github.com/sjtu-sai-agents/X-Master}
}

\begin{document}

\maketitle


\begin{center}
    \vspace{-9mm}
    \begin{tcolorbox}[
        width=1.0\linewidth,  
        colback=bgcolor,
        boxrule=0pt,
        arc=0mm,
        boxsep=0mm
    ]
    \textit{{SciMaster is a series of studies aimed at developing general-purpose scientific AI agents. In Part I, \emph{X-Master} establishes the foundational architecture, laying the groundwork for enhancing the general capabilities of AI agents.}}
    \end{tcolorbox}
\end{center}

\begin{figure}[h]
    \centering
    \vspace{-3mm}
    \includegraphics[width=0.9\linewidth]{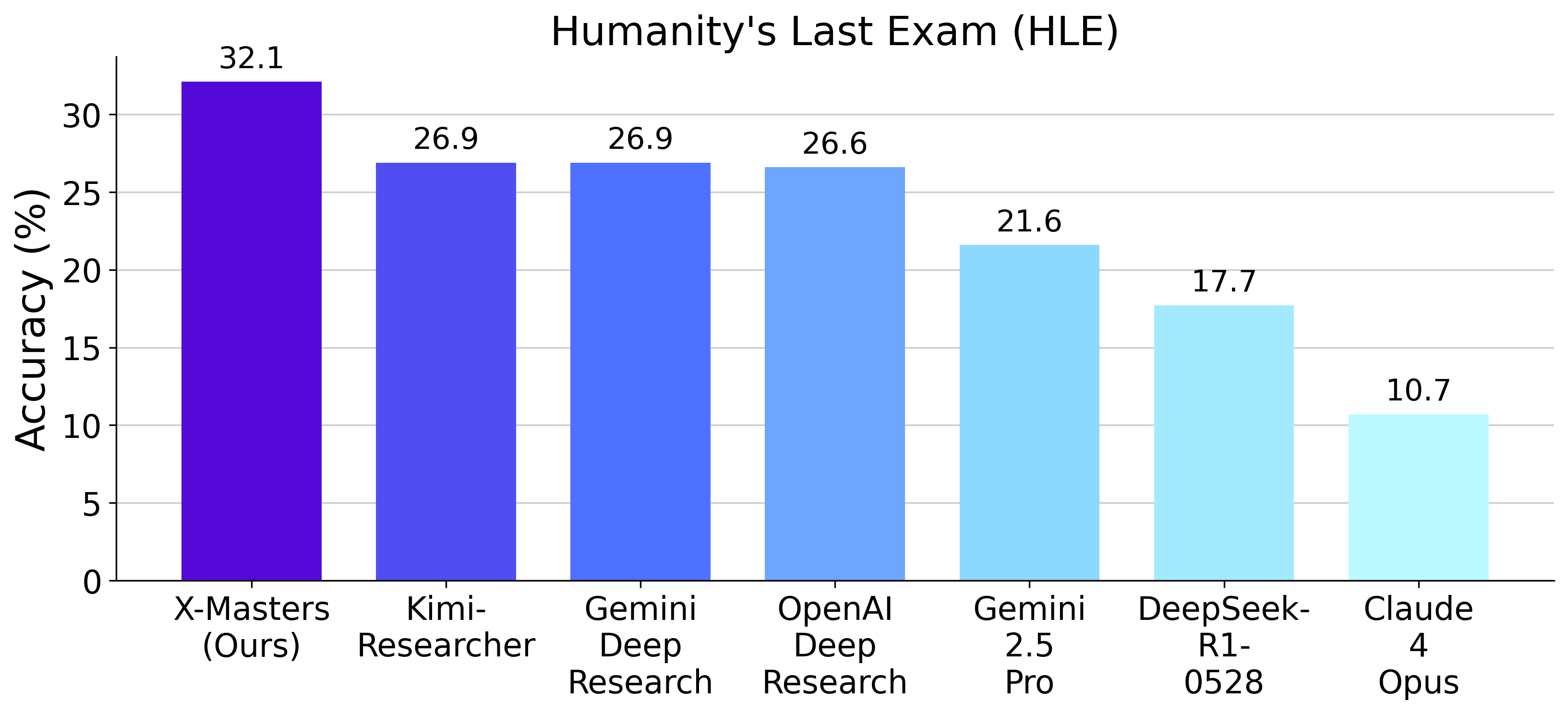}
    \caption{Comparisons on Humanity's Last Exam. Our \emph{X-Masters} achieves the state-of-the-art score of \textbf{32.1\%}, surpassing deep research products from Kimi, Gemini, and OpenAI.}
    \label{fig:main_hle}
\end{figure}

\begin{abstract}
The rapid advancements of AI agents have ignited the long-held ambition of leveraging them to accelerate scientific discovery. Achieving this goal requires a deep understanding of the frontiers of human knowledge. As such, Humanity’s Last Exam (HLE) provides an exceptionally challenging touchstone for evaluating scientific AI agents. In this work, we aim to construct the foundational architecture for general-purpose agents and validate the capabilities through leading performance on HLE. To achieve this, we introduce \emph{X-Master}, a tool-augmented reasoning agent designed to emulate human researchers by interacting flexibly with external tools during its reasoning process. This agent, guided by the conceptualization of code as an interaction language, can flexibly leverage built-in Python libraries and our customized tools to augment the reasoning.
We further scale its capabilities through \emph{X-Masters}, a scattered-and-stacked agentic workflow that systematically enhances breadth and depth of reasoning.
Our open-source solution, \emph{X-Masters}, sets a new state-of-the-art record on HLE with a score of 32.1\%, surpassing OpenAI's and Google's Deep Research (26.6\% and 26.9\%) and becoming the first to exceed the 30\% threshold.
This work allows us to gain a deeper understanding of complex task-solving and accumulates valuable experience that can inform future advancements, guiding subsequent model training.
\end{abstract}

    

\section{Introduction}

Artificial intelligence (AI), particularly in the form of large language models (LLMs)~\citep{chatgpt,dubey2024llama,qwen2.5} is evolving at an unprecedented rate.
In just two years, the landscape has shifted dramatically: from the conversational abilities of models like GPT-3.5/4~\citep{ouyang2022training,openai2023gpt4}, to the emergence of powerful reasoners like DeepSeek R1~\citep{guo2025deepseek}, and now towards AI agents like OpenAI's o3~\citep{o3} and Google's Gemini 2.5 Pro~\citep{gemini-2.5-pro} with flexible tool-use capabilities.
This shift towards general-purpose agents has ignited the long-held ambition of leveraging AI agents to accelerate scientific discovery~\citep{gottweis2025towards}, where they can catalyze breakthroughs by reasoning over vast knowledge and exploring ideas beyond the constraints of human cognition.~\emph{In this series of studies, we aim to progressively build up \emph{SciMaster}, open-source, general-purpose scientific AI agents.}

To enable AI-driven scientific discovery, an essential prerequisite is that an AI agent first demonstrate a deep grasp of the current frontiers of human scientific knowledge.
To this end, Humanity’s Last Exam (HLE)~\citep{hle} serves as a critical and exceptionally challenging touchstone.
Developed through a global collaboration of nearly 1,000 subject experts from over 500 institutions, HLE comprises diverse, expert-level challenges at the frontier of human knowledge.
Consequently, strong performance on HLE is highly indicative of an agent's aptitude for navigating the complex problems inherent in advanced scientific research.
While promising strides have been made on HLE by leading models from OpenAI (26.6\%)~\citep{openai_deep_research} and Google DeepMind (26.9\%)~\citep{gemini-2.5-pro}, their closed-source nature severely limits community understanding and participation, hindering widespread exploration and innovation. 

In light of this, in the first part of this series, we focus on constructing the foundational architecture for general-purpose agents and validating its general capabilities through leading performance on HLE. Our work explores a practical roadmap of inference-time computation that enables open-source models to lead.
This approach, bypassing the need for extensive model training, allows us to gain a deeper understanding of diverse task-solving and accumulates valuable experience that can inform future advancements, guiding subsequent model training.
By openly sharing the insights and methodologies developed throughout this process, we hope to foster greater participation and accelerate progress within the field.

To achieve this goal, we introduce~\emph{X-Master}, a general tool-augmented reasoning agent designed to flexibly interact with external tools during its reasoning process.
As a general-purpose agent architecture, \emph{X-Master} will provide the architectural foundation for the development of \emph{SciMaster}.
The design rationale of \emph{X-Master} is to emulate the dynamic problem-solving process of human researchers who fluidly pivot between internal reasoning and external tool-use.
This creates a symbiotic loop: tool outputs provide crucial feedback to sharpen the agent's reasoning, while clearer reasoning leads to more intelligent and effective tool use.
The core mechanism enabling this process is the conceptualization of code as an interaction language.
When confronted with a problem it cannot solve internally, \emph{X-Master} formulates a precise plan of action as a code block.
This "plan" is then executed, interfacing with any required resource—from the numerical power of NumPy and SciPy to our custom-designed toolkit for live web searches and data extraction.
The result is seamlessly absorbed back into the agent's context, enriching its understanding and informing its subsequent reasoning.
This architecture transforms the model from a static reasoner into an agile problem-solving agent, capable of actively seeking out and utilizing information just as a human would.

To unlock the full potential of our \emph{X-Master}, we design a scattered-and-stacked agentic workflow designed to scale its intelligence at inference time.
This workflow, called~\emph{X-Masters}, is engineered to systematically enhance both the breadth and depth of reasoning by orchestrating a multi-agent cognitive process, where instances of our \emph{X-Master} adopt several specialized roles.
The process begins with the `scattered' phase to establish breadth: multiple Solver agents work in parallel to generate a diverse array of solutions, while Critic agents correct their potential flaws.
Then, the `stacked' phase enhances depth, where Rewriter agents synthesize all preceding outputs into superior solutions before a final Selector agent adjudicates the single best answer.
Through such broad exploration and in-depth improvement, this workflow significantly enhances the capabilities of solving complex problems.

\textbf{Our \emph{X-Masters} sets a new record on Humanity's Last Exam with a remarkable score of 32.1\%.
This score exceeds the previous records held by OpenAI and Google's Deep Research by a substantial margin of 5.5 and 5.2 points respectively, being the first in the world to surpass the 30\% threshold.}
Practically, when faced with a challenging problem and uncertain about which model or method to use, \emph{X-Masters} offers one of the best—if not the best—solutions available. Critically, while these top-performing products remain closed-source, our solution is fully open-sourced, providing all the essential details for replications and further improvements.

In summary, the primary contribution of this paper is not a single, novel method, but rather the sharing of valuable "know-how" with the research community. We demonstrate the possibility of achieving leading performance on an important benchmark, using accessible open-source models. 
In the future, we plan to build upon this work by developing an end-to-end trained system that internalizes these capabilities. 
Related code will be publicly available to encourage further research and collaboration.

\begin{figure}[t]
    \centering
    \includegraphics[width=1.0\linewidth]{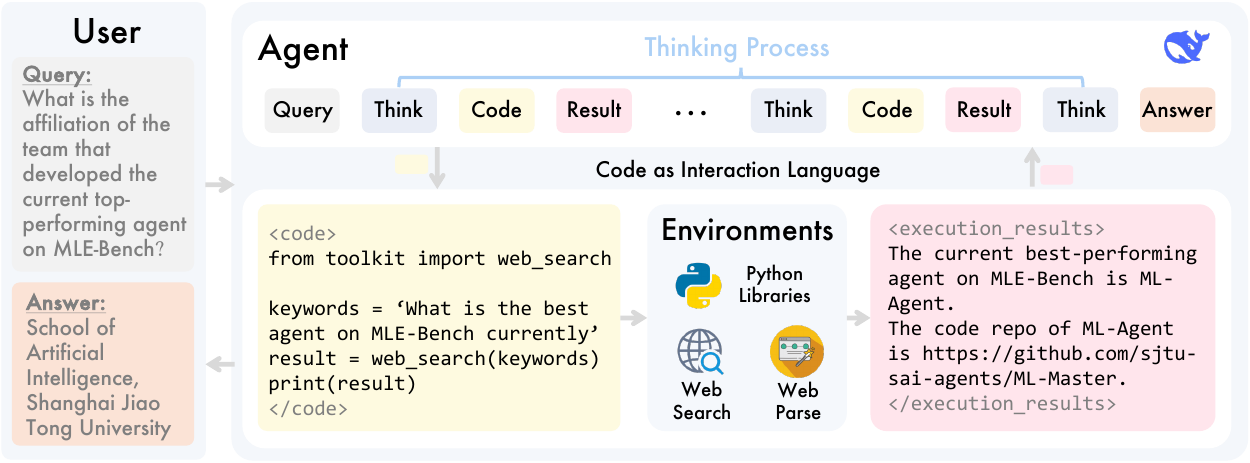}
    \caption{Overview of our \emph{X-Master}, a tool-augmented reasoning agent. Given a user query, the agent starts the thinking process, where interaction with the environments such as tool calling would be invoked by generating a code snippet. The execution results will be appended to the agent's context, enriching its understanding and informing its subsequent thinking. In this case, the agent invokes interactions for three times (search to get the GitHub link, parse to get the arXiv paper link, then parse to get the affiliation) until it arrives at the final answer.}
    \label{fig:overivew}
\end{figure}

\section{Tool-Augmented Reasoning Models as Agents} \label{sec:agent}

This section introduces our tool-augmented reasoning agent, starting from an overview, to our core design of code as interaction language, to tools for interaction with the internet.

\subsection{Overview}

\emph{X-Master} is a tool-augmented reasoning agent powered by open-source models (e.g., DeepSeek-R1~\citep{guo2025deepseek}) designed for flexible interaction with external tools during its deep thinking process.
To achieve this without requiring extensive model retraining, our approach primarily centers on the dynamic manipulation of the model's context.

Given a query from the user that is potentially time-sensitive and knowledge-intensive, the reasoning agent initiates a thought process akin to that of non-agentic models.
Distinctively, our method guides this agent to engage with external environments (e.g., the internet) as needed, for instance, when real-time information is required.
Specifically, if the agent intends to interact with the environments, the agent is guided to generate \colorbox{code}{Python code} that accurately reflects its current requirements.
This generated code is then extracted and dispatched to a code executor, which provides access to various Python libraries and our implemented tools.
The \colorbox{result}{execution results} are subsequently appended to the model's context, enabling the reasoning model to seamlessly continue its thought process.
Notably, the agent possesses the capability to initiate such interactions iteratively, either to acquire novel information or to validate its existing conclusions.
Please refer to Figure~\ref{fig:overivew} for an overview.

This methodology effectively augments the reasoning capabilities of models by extending their access to external environments, thereby transcending the limitations of their inherent knowledge and intelligence boundaries.
In the following, we introduce how the agent interacts with environments via coding, how a non-agentic model is guided towards an agent, and our principal tools.

\subsection{Code as Interaction Language}

Our \emph{X-Master} is designed to generate Python code as a language to interact with external environments.
Specifically, during the thinking process of the agent (i.e., between \thinkStart and \thinkEnd), the agent could either generate non-code texts for reasoning or codes within the special tokens of \codeStart and \codeEnd for interacting with the environments.
Once this pattern is detected via string matching, the code within would be extracted and executed in a sandbox environment, where various python libraries and tools are accessible.
The executed result is then appended to the model's context, enclosed by two special tokens \resultStart and \resultEnd.
Subsequently, the reasoning model continues the thinking process, interpreting the execution results and reasoning further, until the next interaction is invoked or the ending of thinking.

Representing interaction intention as Python code offers three advantages:
(1) Universality: almost all functions can be implemented through writing code, facilitating the agent to address various tasks.
(2) Accuracy: coding language can reflect the accurate needs of agents in a compact and logical manner, enabling efficient interaction.
(3) Compatibility: using Python code indicates that the reasoning models can access any existing well-established libraries from the community.

\subsection{Initial Reasoning Guidance}

Since currently available strong reasoning models (e.g., DeepSeek-R1) are inherently non-agentic and often exhibit limited instruction-following capabilities, conventional prompt engineering alone proves insufficient to reliably guide these models towards expected agentic behavior.
Instead, we introduce a simple yet effective mechanism: Initial Reasoning Guidance.

Specifically, rather than allowing the reasoning models to commence their unconstrained thinking process immediately upon receiving a user query, we embed a series of guiding texts directly after the model's initial \thinkStart token.
These guiding texts are deliberately crafted from the perspective of the reasoning agent itself, speaking in the first person. For instance, such guidance includes statements like:
(1) `I can answer this query effectively by leveraging access to external environments.'
(2) `Every time I determine the need for interaction with external tools, I will generate Python code enclosed between \codeStart and \codeEnd tags.'

By concatenating these meticulously designed self-statements into the model's context, we effectively lead the model to "believe" in its own enhanced capabilities.
Consequently, the model is empowered to spontaneously generate and execute code, interact with its environment, and ultimately function as a capable agent, even without explicit finetuning for agentic behavior.

\subsection{Tools}

By expressing interactions as Python code, our \emph{X-Master} gains strong flexibility, capable of integrating a wide spectrum of tools—from existing Python libraries to custom-built modules, and even generating new tools dynamically. Notably, when we equip \emph{X-Master} with a suite of professional scientific computing tools, empowering it to solve domain-specific research problems and advance scientific discovery. Here, we mainly introduce our custom-built tools for information-seeking tasks.

To emulate human-like online information-seeking behavior, we design two core tools: web search and web parse. The web search tool facilitates the agent's ability to identify relevant webpages based on the question.
It provides concise summaries for each retrieved page, enabling the agent to strategically determine which links warrant deeper exploration.
The web parse tool is employed when the agent requires a more thorough examination of a selected webpage to extract information directly related to the user query.

\textbf{Web search.}
The web search tool leverages the Google search engine to identify the most relevant webpages for a given question.
It furnishes three types of highly valuable information:
(i) Entity-related facts: 
For questions containing identifiable entities (e.g., a company or application), the tool can detect these and retrieve structured facts from its knowledge graph, including name, description, and key attributes.
Extracting these facts enables the agent to rapidly comprehend the core concept of the question, thereby providing crucial context for subsequent reasoning.
(ii) Relevant webpage previews:
For each pertinent page, the tool presents a preview comprising its title, URL, and a descriptive snippet.
This functionality assists the agent in quickly discerning the page's content and prioritizing those warranting deeper exploration.
(iii) Related search queries: The tool also provides common related queries, which offer the agent avenues for follow-up searches and contribute to a broader understanding of the topic.

\textbf{Web parse.}
The web parse tool offers support for two distinct parsing strategies, tailored for general webpages and scientific papers, respectively:
(i) General webpage parsing: 
This strategy initiates by extracting the primary content from the target webpage. 
To ensure robust operation, a fallback mechanism is incorporated to manage instances where direct content extraction may encounter failures. 
Following content acquisition, the tool proceeds to identify segments highly relevant to the query. 
Furthermore, it automatically detects and returns links along with brief descriptions of pertinent subpages. 
This feature enables the agent to explore information more deeply, effectively emulating human behavior of navigating interconnected web content to gain comprehensive topic understanding.
(ii) Scientific paper parsing: For scientific papers, the tool employs a two-step strategy to guarantee reliable content retrieval. 
Initially, it endeavors to retrieve an HTML version of the publication from ar5iv.
In the event of an unsuccessful or incomplete HTML fetch, the system transparently reverts to downloading the authoritative PDF document.
Upon successful acquisition of the paper's full content, the tool automates the process of extracting information directly correlated with the question.

In this way, the web search and web parse tools not only help the agent find relevant information, but also encourage the agent to explore the web in a more human-like way—by searching, scanning, clicking, and digging deeper as needed.

\section{Scattered-and-Stacked Agentic Workflows for Inference-Time Computation}
To further scale the potential of our powerful tool-augmented reasoning agent at inference time, we propose \emph{X-Masters}, an agentic workflow based on the principles of scattered and stacked process.
Scattering enables broad problem-solving across multiple instances, generating diverse solutions, while stacking facilitates iterative improvements through rewriting and final selection.
The agent, powered by \emph{X-Master}, performs four specialized roles: Solver, Critic, Rewriter, and Selector in a tool-integrated manner, to execute this scattered-and-stacked process.

\begin{figure}[t]
\centering
\includegraphics[width=0.96\textwidth]{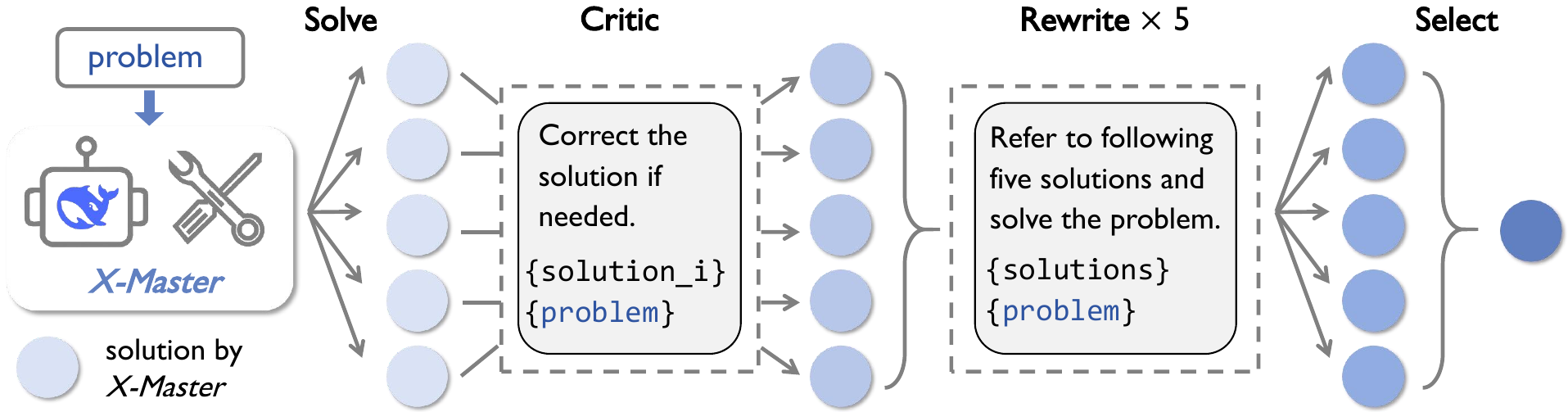}
\caption{Overview of our \emph{X-Masters}, a scattered-and-stacked agentic workflow. The workflow leverages \emph{X-Master} as different roles to enhance solution quality at inference. It includes (1) Solvers generating five initial solutions, (2) Critics refining the initial solution, (3) Rewriters synthesizing all five solutions to generate new five, and (4) Selector choosing the best solution.}\label{fig:workflow}
\end{figure}

\subsection{Workflow Overview}

As depicted in Figure~\ref{fig:workflow}, this workflow alternates between expansive exploration (scattering) and sequential refinement (stacking), leveraging \emph{X-Master} at every stage.
Specifically, the workflow includes four stages: (1) broad initial solutions generation by Solver, (2) solution refinement by Critic, (3) broad solutions rewriting, and (4) final selection.
In Stage 1, Solver generates five initial solutions using tool-augmented reasoning for exploring scattered potential solutions.
In Stage 2, Critic diagnoses flaws and amends the solution.
Stage 3 stacks these results with Rewriter referring to all solutions and rewriting five.
Stage 4 stacks once more: the Selector compares all five solutions from Rewriters, then selects the optimal one. 

Overall, scattering aims to explore different diverse solutions, while stacking enhances the quality by integrating the intelligence of previous solutions. 
The scattered–stacked design combines breadth and depth: broadening solutions fosters diversity, whereas iterative improvement enhances precision.

\subsection{Scattering for Broad Exploration}
The scattering phase of the workflow involves two distinct broad processes to explore diverse solutions. For breadth, the agent answers the query five times concurrently, each with the same temperature, leveraging the random nature of the decoding process of large language models. In Stage 1, the Solver generates five initial solutions concurrently, each with tool-augmented reasoning. Afterward, the Critic concurrently evaluates each of these five solutions, identifying flaws and providing corrections or improvements, ensuring that all generated solutions are internally consistent and refined.
The second part of the scattering phase (Stage 3) enhances the quality of solutions by the rewriting process. The Rewriter generates five refined versions of each initial solution. This ensures that multiple improvements are explored simultaneously, creating a diverse set of reworked answers. By combining breadth in both solution generation and rewriting, the system enhances its ability to explore and refine multiple solution paths, providing a richer set of potential solutions for further refinement.

\subsection{Stacking for Sequential Refinement}
The stacking phase enhances and combines solutions through rewriting and selection. In Stage 3, the Rewriter synthesizes the scattered solutions into a coherent draft, resolving redundancies and contradictions. Then, a second scattering occurs, where five Rewriters comprehensively analyze the previous five solutions and generate five. Finally in Stage 4, the Selector aggregates these refined solutions, selecting the most optimal one based on logical consistency and factual accuracy. By treating selection as a form of stacking, the workflow ensures that the final output combines diverse insights and iterative refinements, resulting in a high-quality solution.

\subsection{Discussion: An Analogy to Rollouts in Reinforcement Learning}

Our scattered-and-stacked architecture is fundamentally a strategy for structured exploration and exploitation, drawing a strong parallel to the concept of "rollouts" in reinforcement learning (RL)~\citep{tesauro1995temporal,sutton1998reinforcement,auer2002finite,guo2025deepseek}.
This underlying logic also resonates with recent advancements in RL, where enabling models to explore diverse solutions and leveraging self-generated rewards has proven effective in training stronger models~\citep{zhao2025absolute,zhao2025learning,prabhudesai2025maximizing}.

(1) The "scattered" phase mirrors the exploratory principle of rollouts in RL that simulate several future trajectories to evaluate the potential of different actions.
Leveraging the agent's (model's) stochastic decoding, each agent explores a different reasoning path, ensuring a diversity of initial strategies and preventing premature convergence on a single, potentially suboptimal idea.
(2) The subsequent "stacked" phase is analogous to the aggregation and "exploitation" step that follows rollouts in RL.
Agents aggregate insights from all parallel explorations to construct a superior, more robust solution, ensuring the final output synthesizes the best elements discovered.

In essence, the scattered-and-stacked workflow operationalizes the exploration-exploitation paradigm at inference time. 
Scattering broadly explores possibilities, while stacking deeply refines them.
This structured thinking allows the system to tackle complex problems more effectively than a single reasoning process, thereby scaling the agent's intelligence.

\section{Experiments}
\subsection{Experimental Setups}
\textbf{Agent.}
We use DeepSeek-R1-0528~\citep{deepseek-r1-0528} as the reasoning model to drive the agent.
The maximum completion of tokens is set to 64k with a temperature of 0.6.

\textbf{Testing.}
Since the currently leveraged model is not multi-modal, we focus on the text-only subset from Humanity's Last Exam~\citep{hle}, which comprises 2,518 samples following~\citep{kimi_researcher}.
For evaluation, we run the workflow for three times and report the average score, and utilize o3-mini~\citep{o3-mini} as the judge model following the official setup.

\textbf{Baselines.} We compare our workflow's performance against systems spanning two categories: research agents (OpenAI's Deep Research~\citep{openai_deep_research}, Google DeepMind's Deep Research~\citep{gemini-2.5-pro}, Kimi-Researcher~\citep{kimi_researcher}) and advanced models (Gemini 2.5 Pro~\citep{gemini-2.5-pro}, DeepSeek-R1-0528~\citep{deepseek-r1-0528}, and Claude 4 Opus~\citep{claude4}).
The results on HLE of these models are taken from existing leaderboards.

\subsection{Main Results on Humanity's Last Exam}

\begin{figure}[t]
    \centering
    \begin{minipage}{0.66\linewidth}
    \includegraphics[width=1\linewidth]{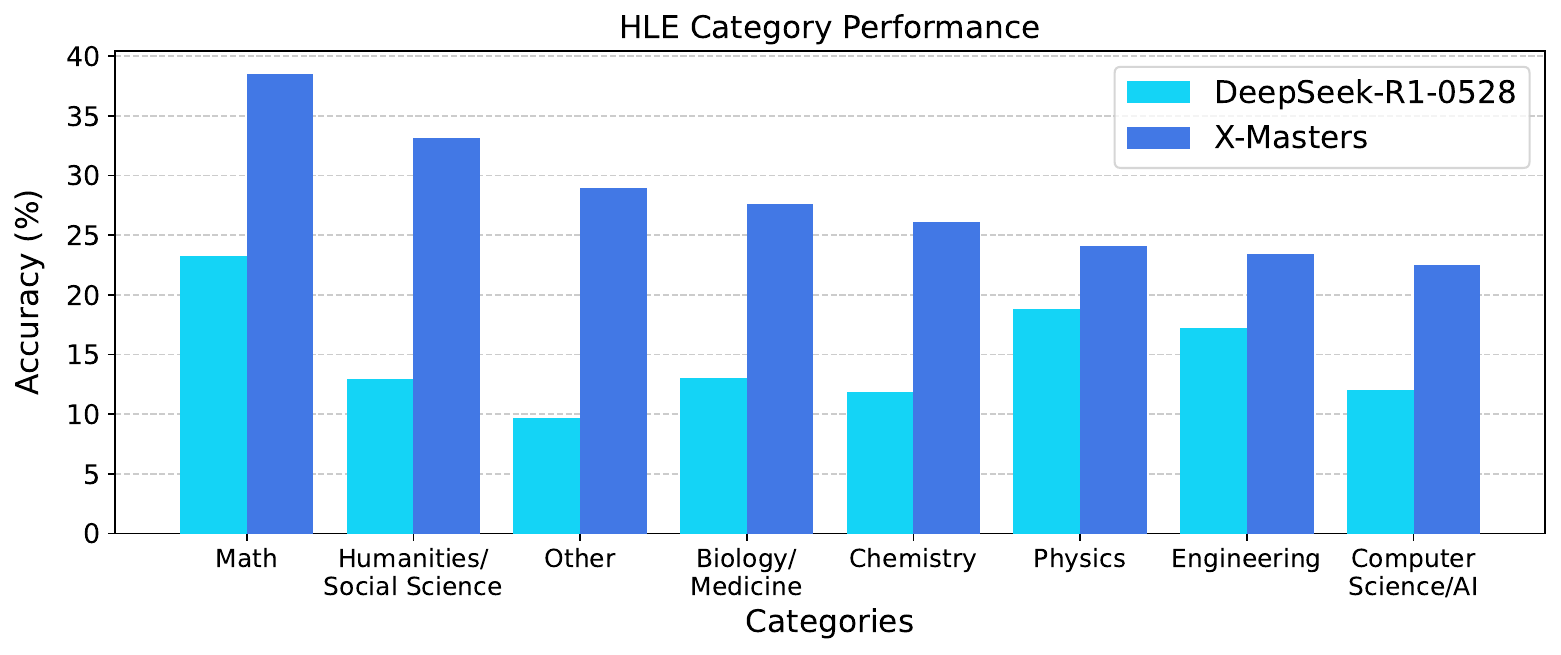}
    \caption{Performance across categories of DeepSeek-R1-0528 and \emph{X-Masters} on HLE.}
    \label{fig:category_dis}
    \end{minipage}
    \hfill
    \begin{minipage}{0.32\linewidth}
    \includegraphics[width=0.968\linewidth]{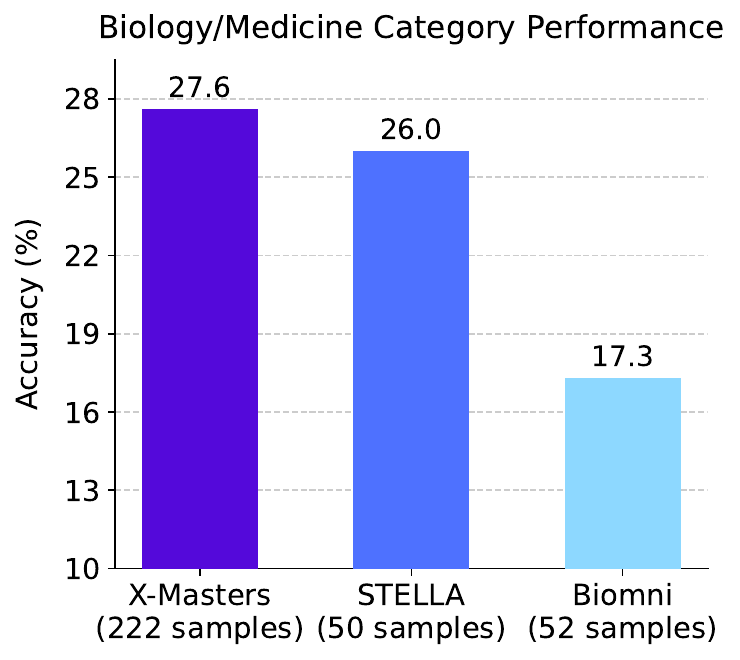}
    \caption{Performance of Biology/Medicine category of HLE.}\label{fig:bio_compare}
    \end{minipage}
\end{figure}
The performance results of \emph{X-Masters} against other systems in Humanity's Last Exam (HLE) are shown in Figure~\ref{fig:main_hle}.
\emph{X-Masters} achieves the highest score of \textbf{32.1\%}, surpassing all existing agents and models, including Gemini Deep Research, and OpenAI Deep Research, which scored 26.9\% and 26.6\%.
Excitingly, this is the first time that a system performs above 30\% on HLE, setting a promising new record.
The leading performance on extremely challenging HLE speaks volumes about the excellence of tool-augmented \emph{X-Master} and the inference-time agentic workflow.  In addition, we report the performance across different categories in Figure~\ref{fig:category_dis}, where \emph{X-Masters} shows a notable improvement across all categories compared to DeepSeek-R1-0528. This highlights the broad applicability and effectiveness of our tool-augmented reasoning and agentic workflow.

We also present the ablation results across workflow stages in Table~\ref{tab:stage_result}. The results reveal progressive gains in the agentic workflow: tool-augmented reasoning (Solver) lifts baseline accuracy by 3.4\%, iterative refinement (Critic and Rewriter) adds 9.5\%, and final selection (Selector) achieves the record 32.1\%. Detailed role-wise analysis follows in Section~\ref{sec:analysis}. 

\subsection{Comparisons with Scientific Agents on Biology}
\textbf{\emph{X-Masters} outperforms existing systems on the Biology/Medicine category of HLE.} Recent works, such as Biomni~\citep{biomni} and STELLA~\citep{stella} have made progress in addressing biological challenges by leveraging LLM agents with an extensive collection of specialized tools. As illustrated in Figure~\ref{fig:bio_compare}, while Biomni gets 17.3\% and STELLA gets approximately 26\%, our \emph{X-Masters} achieves 27.6\% accuracy in Biology/Medicine category. Note that these two systems evaluate on their selected questions from the Biology/Medicine category; in contrast, we evaluate \emph{X-Masters} on the complete 222 text-only questions. The domain-specific results illustrate~\emph{X-Masters}' advanced capabilities in solving complex biomedical problems, emphasizing its significant potential for contributing to biological research.

\textbf{\emph{X-Masters} achieves the state-of-the-art performance on the biology benchmark TRQA.}
Beyond general scientific challenges, we evaluate \emph{X-Master} on TRQA-lit (choice)~\citep{trqa}, a specialized benchmark comprising 172 multiple-choice questions on biological research. The TRQA-lit dataset targets complex research tasks in the biological domain, including the identification of therapeutic targets, and biomedical mechanisms. These tasks represent high-level challenges faced by human experts in the field.

\begin{figure}[t]
\centering
\begin{minipage}{0.56\linewidth}
    \centering
    \includegraphics[width=0.99\textwidth]{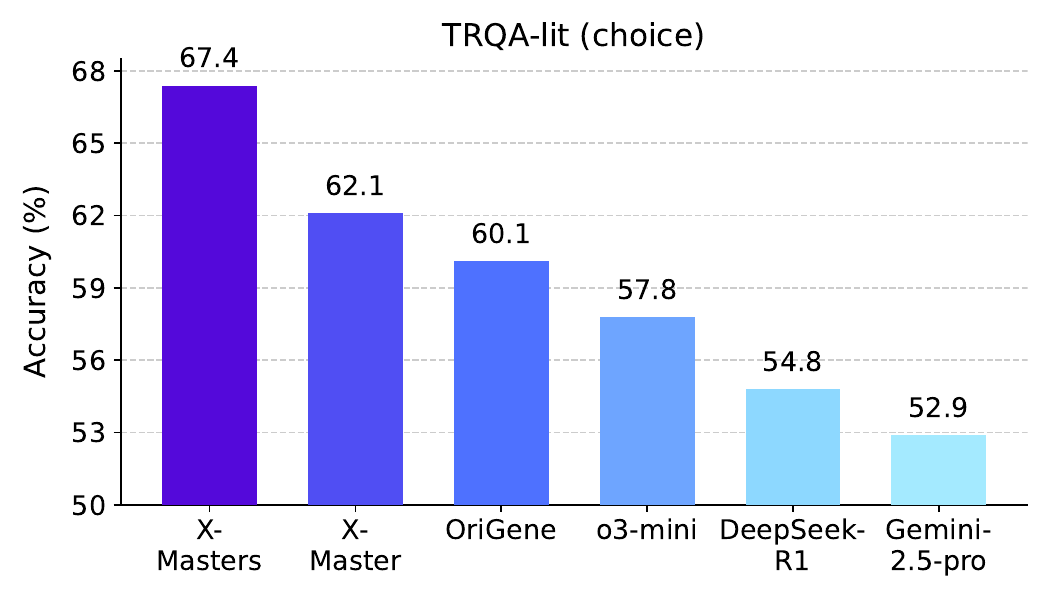}
    \caption{Performance of \emph{X-Masters} with other models on a biology benchmark TRQA-lit (choice). Without any modification, \emph{X-Masters} also achieves the state-of-the-art performance on this benchmark.}\label{fig:trqa}
\end{minipage}
\hfill
\begin{minipage}{0.41\linewidth}
    \centering
    \includegraphics[width=0.98\textwidth]{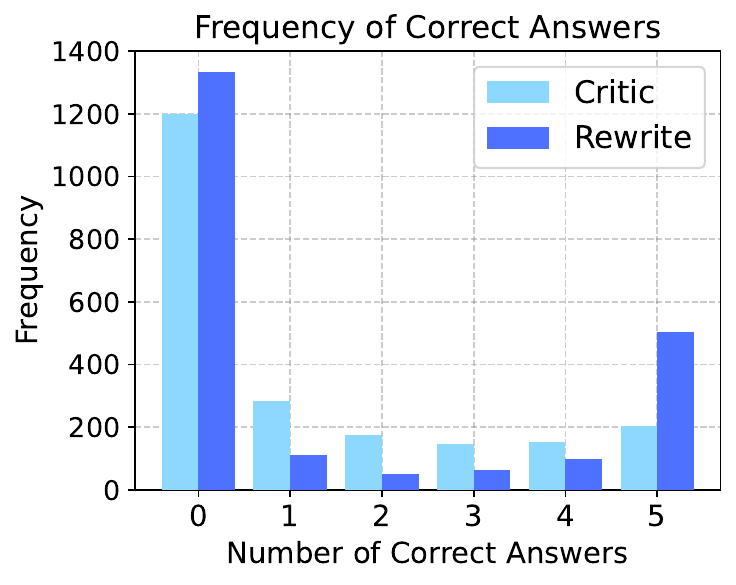}
    \caption{Frequency of correct answers before and after rewriting stage. Rewriting significantly enhances the likelihood of all 5 solutions being correct.}\label{fig:rewrite_dis}
\end{minipage}
\end{figure}

Here, we report the results of \emph{X-Master} and \emph{X-Masters} in Figure~\ref{fig:trqa}, where the results of other models are referred from paper~\citep{trqa}. 
Results show that (1) standalone \emph{X-Master} achieves 62.1\%, already outperforming other models with tool-augmented reasoning. (2) With the simple agentic workflow, \emph{X-Masters} achieves a state-of-the-art \textbf{67.4\%}, demonstrating the effectiveness of scattered exploration and stacked selection. (3) Compared with OriGene~\citep{trqa}, a multi-agent system that integrates over 500 expert tools, our \emph{X-Master}, utilizing only two web tools, obtains higher accuracy. This reinforces the demonstrated efficiency of \emph{X-Master}’s tool-augmented reasoning process, where broad exploration and stacked selection enable it to effectively solve complex biological tasks.

\subsection{Analysis} \label{sec:analysis}

\begin{table}[tb]
    \centering
    \caption{Progressive accuracy improvement across stages of the \emph{X-Masters} agentic workflow versus the DeepSeek-R1-0528. Tool augmentation and inference-time computation drive performance gains.}

    \begin{tabular}{c|ccccc|c}
    \toprule
    Method & R1 & Solver & Critic & Rewriter & Selector & Accuracy (\%) \\ \midrule
    Reasoning Model without Tool & \checkmark & \ding{55} & \ding{55} & \ding{55} & \ding{55} & 17.7 \\
    \rowcolor{blue!5} & \checkmark & \checkmark & \ding{55} & \ding{55} & \ding{55} & 21.1 \\
    \rowcolor{blue!5}&\checkmark & \checkmark & \checkmark & \ding{55} & \ding{55} & 25.0 \\
    \rowcolor{blue!5}&\checkmark & \checkmark & \checkmark & \checkmark & \ding{55} & 30.6 \\
    \rowcolor{blue!5}\multirow{-4}{*}{\textbf{\emph{X-Master} (Ours)}} &\checkmark & \checkmark & \checkmark & \checkmark & \checkmark & \textbf{32.1} \\ \bottomrule
\end{tabular}
    \label{tab:stage_result}
\end{table}

\textbf{Tool-augmented reasoning enhances DeepSeek-R1.} In our agentic workflow, Solver accesses external tools an average of 3 times per query while generating initial solutions. For simplicity, we use the average score of the parallel generated initial 5 solutions to estimate the pass@1. Shown in Table~\ref{tab:stage_result}, when tool support is enabled, pass@1 rises from 17.7\% to \textbf{21.1\%}, indicating the effectiveness of inference-time tool augmentation in improving the accuracy of the first attempt. 

\textbf{Critic boosts refinement of solution.} In the scattered solution exploration, each solution is first generated by the Solver and then refined by the Critic.
We compare the average scores of the initial five solver solutions before and after applying the criticism in Table~\ref{tab:stage_result}. We find that this stage delivers substantial improvement: the average accuracy rises from 21.1\% to \textbf{25.0\%}. This demonstrates the Critic's crucial role in identifying errors and enhancing solution quality.

\textbf{Rewriting significantly improves solution quality.} From Table~\ref{tab:stage_result}, the Rewriter consolidates insights from scattered solutions, achieving a 5.6 point leap to \textbf{30.6\%} accuracy. To further understand the essence of the rewriting stage, we compare the frequency of the number of correct answers among 5 solutions before and after applying the rewriting. Shown in Figure~\ref{fig:rewrite_dis}, we can observe a noticeable increase in the frequency of higher correctness numbers ($\geq3$), particularly for solutions with 5 fully correct answers. This shift suggests that the rewriting process effectively improves the quality of the solutions. In addition, the movement from mid-frequency correctness to higher correctness also makes the subsequent Selector choose the optimal solution from 5 solutions. Therefore, the rewrite phase serves as a critical quality amplifier, transforming scattered candidate solutions into a refined, high-value solution pool.

\begin{wraptable}{r}{5cm}
    \centering
    \caption{Ablation study of scattering and stacking feature in the agentic workflow.}
    \tabcolsep=0.2cm
    \begin{tabular}{cc|c}
    \toprule
        scatter & stack & Accuracy (\%) \\ \midrule
        \ding{55} & \checkmark & 25.5 \\
        \checkmark & \ding{55} & 25.0 \\
        \rowcolor{blue!5} \checkmark & \checkmark & 32.1 \\ \bottomrule
    \end{tabular}
    \label{tab:ablation_scatter}
\end{wraptable}
\textbf{Scattering and stacking matter in the agentic workflow.} We conduct an ablation study of our scattered-and-stacked agentic workflow, shown in Table~\ref{tab:ablation_scatter}. When no scattering, Solver only generates one solution, Critic refines the solution, and Rewriter analyzes the single solution to generate a new one. The absence of parallel exploration slashes accuracy to 25.5\%. When no stacking, rewriting and final selection stages are removed, the workflow loses its depth-wise improvements and accuracy drops to 25.0\%. Enabling both scattering and stacking boosts performance to 32.1\%. The results confirm that scattering supplies diverse solutions, while stacking distills them into a higher-quality answer; only their synergy realizes both breadth and depth in reasoning.

\begin{figure}[h]
    \centering
    \includegraphics[width=1.0\linewidth]{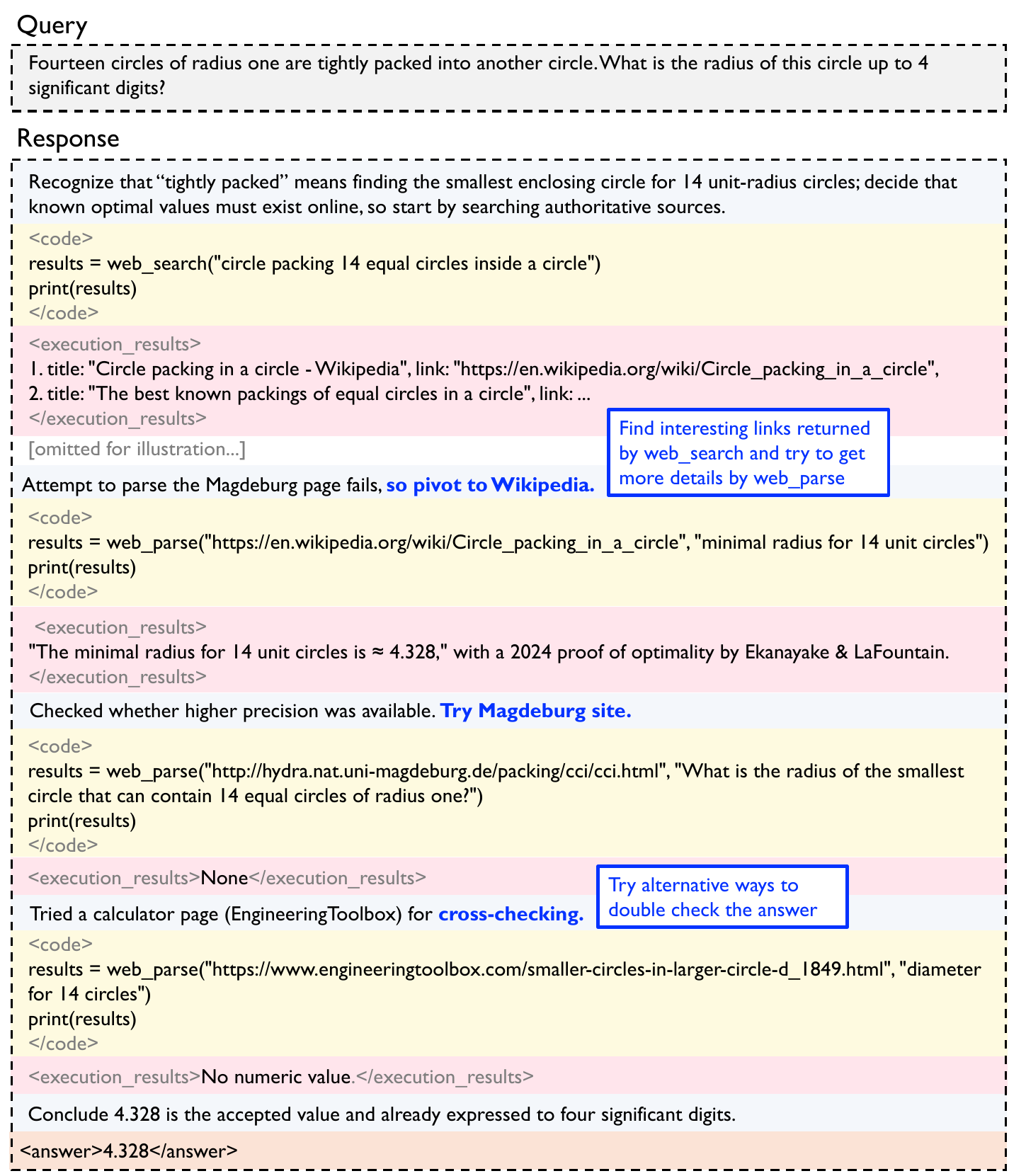}
    \caption{Case 1. \emph{X-Master} fluently handles the case where the tools do not work as expectation by adjusting the usage; and ensures the correctness of answer by cross-checking.}
    \label{fig:case_1}
\end{figure}

\begin{figure}[t]
    \centering
    \includegraphics[width=1.0\linewidth]{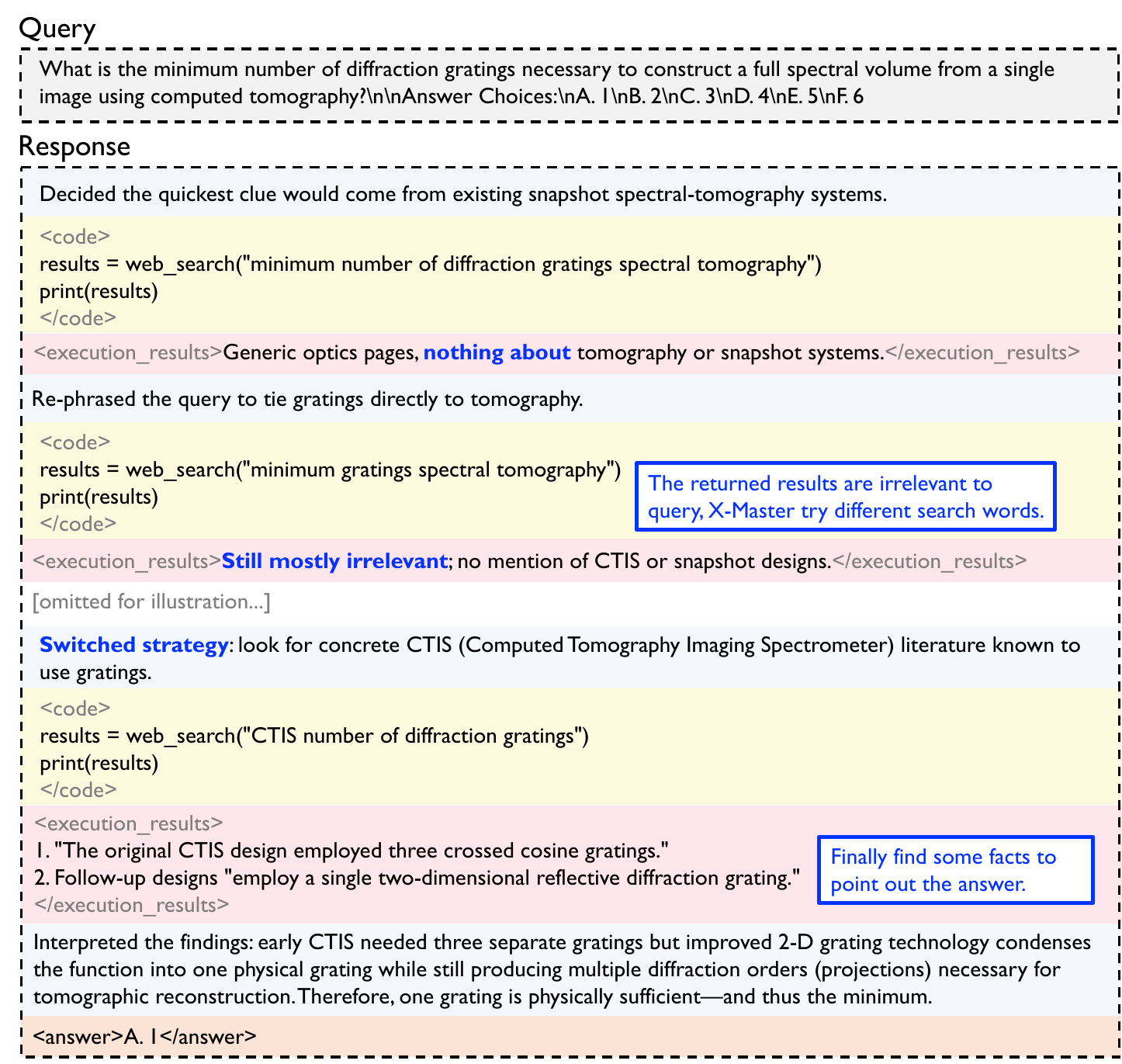}
    \caption{Case 2. \emph{X-Master} fluently handles the case where the tools do not return the expected content by trying different strategies.}
    \label{fig:case_2}
\end{figure}

\begin{figure}[t]
    \centering
    \includegraphics[width=1.0\linewidth]{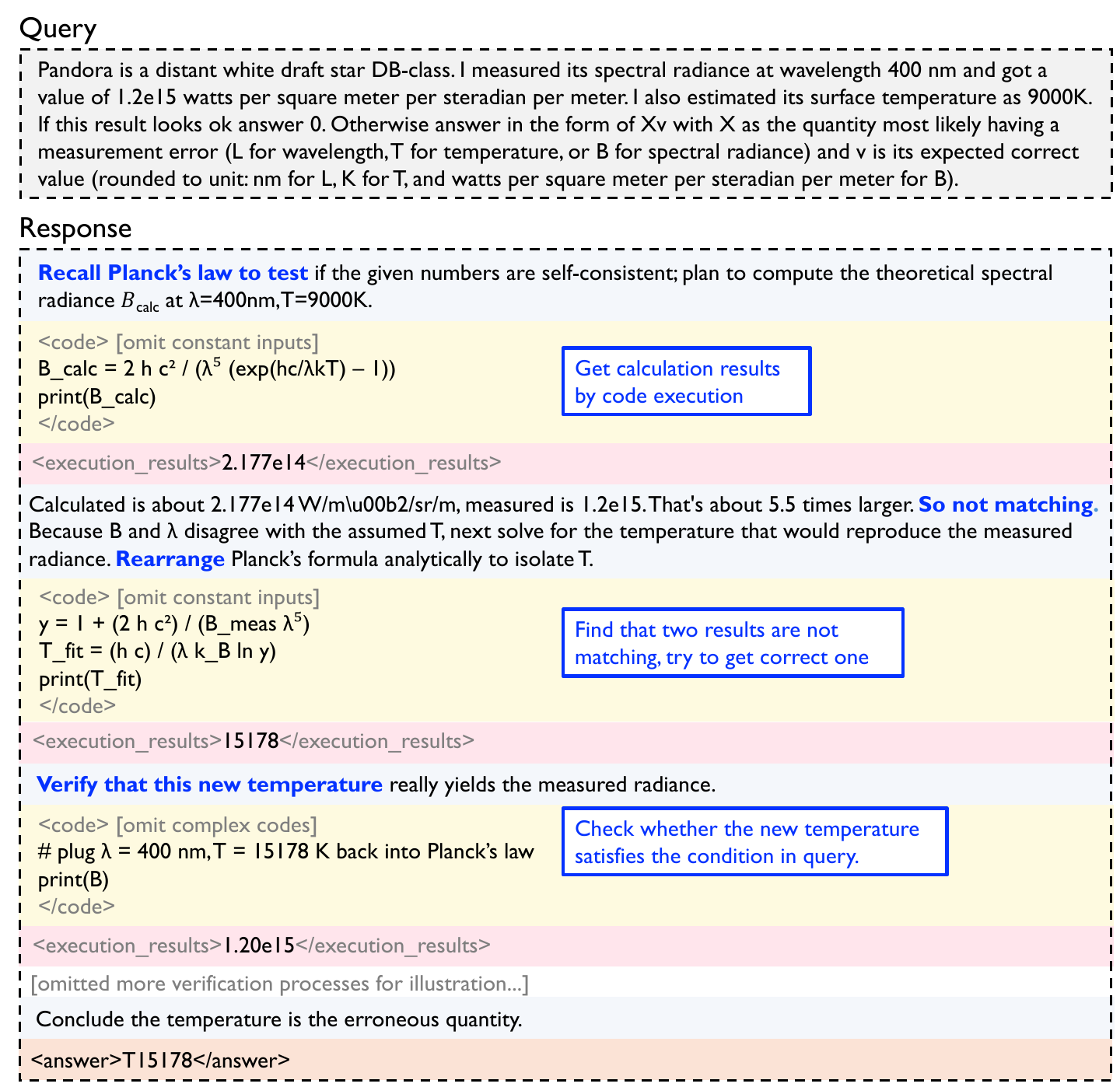}
    \caption{Case 3. \emph{X-Master} conducts computations, adjusts strategies when encountering unmatched results, and verifies the final answer by writing python code.}
    \label{fig:case_3}
\end{figure}

\textbf{Examples.} We provide three examples of \emph{X-Master's} problem-solving trajectories in Figure~\ref{fig:case_1},~\ref{fig:case_2},~\ref{fig:case_3}.

\section{Related Works}

\textbf{Tool-augmented LLMs.}
There are several recent works that focus on leveraging external tools to augment LLMs, which apply different strategies for tool invocation.
\textit{(1) Prompt engineering for structured outputs.}
Some methods, like Agentic Reasoning~\citep{agentic_reasoning}, leverage prompt engineering to guide LLMs in generating outputs in a specific format for subsequent processing by other LLM calls. 
While effective for fixed agentic workflows, these approaches often lack the adaptability needed for complex problems requiring iterative tool usage.
\textit{(2) Special token-guided tool invocation.}
Methods such as Search-R1~\citep{search-r1}, Search-o1~\citep{search-o1}, R1-Searcher~\citep{r1-searcher}, WebThinker~\citep{webthinker}, and WebDancer~\citep{webdancer} guide the model to generate tool invocation content within specific special tokens (e.g., `search').
The results from these tools are then re-inserted into the model's context. 
However, a significant limitation here is that integrating each new tool necessitates manual modifications to the tool invocation rules.
This makes them less practical for diverse and complex real-world tasks, such as scientific research, which often demand a wide array of tools.
\textit{(3) Code generation for computation.}
Methods such as ToRL~\citep{ToRL} and ReTool~\citep{Retool} enable LLMs to generate code for computational tasks, like mathematical calculations.
However, their scope is typically limited to computation, overlooking the broader need for custom tools tailored to real-world scenarios beyond mere calculations.
Addressing the limitations of existing methods, our work offers the most general approach to tool usage currently available.
We leverage Python code as an interaction language, which allows our agent to access: built-in Python libraries (e.g., for computation), our customized tools (e.g., web search), and even build its own tool during inference time.
This unique extensibility empowers our agent to meet the demands of complex real-world tasks.

\textbf{Agentic workflows.}
Agentic workflows empower LLMs to tackle more complex tasks by orchestrating multiple LLM calls and tool interactions.
For instance, AI co-scientist~\citep{gottweis2025towards} leverages multiple agents and tools for scientific research; ChatDev~\citep{qian2024chatdev} and MetaGPT~\citep{hongmetagpt} design agentic workflows for software development; MAS-GPT~\citep{masgpt} generates query-specific workflows represented as Python code.
However, a common limitation in these existing methods is that their agents operate in a fixed single-turn manner.
This means they perform only one action—either generating a textual response or calling a tool—for each assigned (sub-)task.
In contrast, our \emph{X-Masters} framework features agents that work in a flexible multi-turn manner.
This allows them to flexibly and iteratively interplay between internal reasoning and external tool usage, enabling a more dynamic and adaptive approach to problem-solving.

\section{Conclusions and Future Works}

This work presents \emph{X-Master}, the foundational agent introduced as Part I of our \emph{SciMaster} series, which aims to enhance the general capabilities of AI agents and accelerate the pace of future scientific discovery. \emph{X-Master} is a general tool-augmented reasoning agent, designed to emulate human problem-solving by flexibly interplaying between internal reasoning and external tool usage.
Our conceptualization of code as an interaction language allows \emph{X-Master} to flexibly interact with diverse resources.
To further scale its intelligence at inference time, we develop a scattered-and-stacked agentic workflow \emph{X-Masters}, systematically enhancing both the breadth of exploration and the depth of exploitation during task solving.
Remarkably, our \emph{X-Masters} achieves a score of 32.1\% on Humanity's Last Exam (HLE), setting a new world record that crosses the 30\% threshold and surpassing the previous leading closed-source models from OpenAI (26.6\%) and Google DeepMind (26.9\%) by substantial margins.

Our primary contribution lies not in proposing yet another algorithm, but in openly sharing the practical "know-how" that enables open-source models to attain—and even surpass—state-of-the-art performance on challenging benchmarks, such as HLE.  By providing a practical roadmap for enhancing LLM capabilities without requiring extensive retraining, we share insights into complex problem-solving and lay the groundwork for future advancements. With this work, we hope to demonstrate to the community that cutting-edge benchmarks are not the exclusive domain of resource-rich industry labs.

\textbf{What's next?}
Looking ahead, \emph{SciMaster} will significantly expand on the architectural foundation of \emph{X-Master}. Our roadmap includes the development of specialized scientific agents and tools to support literature analysis, scientific computing, and experimental workflows. In addition, we aim to build end-to-end trained agents that fully internalize the sophisticated reasoning and tool-use capabilities showcased by \emph{X-Masters}.
{
\small
\bibliographystyle{plainnat}
\bibliography{ref}

\begin{thebibliography}{35}
\providecommand{\natexlab}[1]{#1}
\providecommand{\url}[1]{\texttt{#1}}
\expandafter\ifx\csname urlstyle\endcsname\relax
  \providecommand{\doi}[1]{doi: #1}\else
  \providecommand{\doi}{doi: \begingroup \urlstyle{rm}\Url}\fi

\bibitem[Anthropic(2025)]{claude4}
Anthropic.
\newblock Introducing claude 4.
\newblock \url{https://www.anthropic.com/news/claude-4}, 2025.
\newblock Accessed: 2025-07-03.

\bibitem[Auer et~al.(2002)Auer, Cesa-Bianchi, and Fischer]{auer2002finite}
Peter Auer, Nicolo Cesa-Bianchi, and Paul Fischer.
\newblock Finite-time analysis of the multiarmed bandit problem.
\newblock \emph{Machine learning}, 47:\penalty0 235--256, 2002.

\bibitem[DeepMind(2025)]{gemini-2.5-pro}
Google DeepMind.
\newblock Gemini 2.5: Our most intelligent ai model.
\newblock \url{https://blog.google/technology/google-deepmind/gemini-model-thinking-updates-march-2025/#gemini-2-5-thinking}, 2025.
\newblock Accessed: 2025-06-24.

\bibitem[DeepSeek-AI(2025)]{deepseek-r1-0528}
DeepSeek-AI.
\newblock Deepseek-r1-0528.
\newblock \url{https://huggingface.co/deepseek-ai/DeepSeek-R1-0528}, 2025.
\newblock Accessed: 2025-06-27.

\bibitem[Dubey et~al.(2024)Dubey, Jauhri, Pandey, Kadian, Al-Dahle, Letman, Mathur, Schelten, Yang, Fan, et~al.]{dubey2024llama}
Abhimanyu Dubey, Abhinav Jauhri, Abhinav Pandey, Abhishek Kadian, Ahmad Al-Dahle, Aiesha Letman, Akhil Mathur, Alan Schelten, Amy Yang, Angela Fan, et~al.
\newblock The llama 3 herd of models.
\newblock \emph{arXiv preprint arXiv:2407.21783}, 2024.

\bibitem[Feng et~al.(2025)Feng, Huang, Qu, Zhang, Qin, Zhong, Jiang, Chi, and Zhong]{Retool}
Jiazhan Feng, Shijue Huang, Xingwei Qu, Ge~Zhang, Yujia Qin, Baoquan Zhong, Chengquan Jiang, Jinxin Chi, and Wanjun Zhong.
\newblock Retool: Reinforcement learning for strategic tool use in llms.
\newblock \emph{arXiv preprint arXiv:2504.11536}, 2025.

\bibitem[Gottweis et~al.(2025)Gottweis, Weng, Daryin, Tu, Palepu, Sirkovic, Myaskovsky, Weissenberger, Rong, Tanno, et~al.]{gottweis2025towards}
Juraj Gottweis, Wei-Hung Weng, Alexander Daryin, Tao Tu, Anil Palepu, Petar Sirkovic, Artiom Myaskovsky, Felix Weissenberger, Keran Rong, Ryutaro Tanno, et~al.
\newblock Towards an ai co-scientist.
\newblock \emph{arXiv preprint arXiv:2502.18864}, 2025.

\bibitem[Guo et~al.(2025)Guo, Yang, Zhang, Song, Zhang, Xu, Zhu, Ma, Wang, Bi, et~al.]{guo2025deepseek}
Daya Guo, Dejian Yang, Haowei Zhang, Junxiao Song, Ruoyu Zhang, Runxin Xu, Qihao Zhu, Shirong Ma, Peiyi Wang, Xiao Bi, et~al.
\newblock Deepseek-r1: Incentivizing reasoning capability in llms via reinforcement learning.
\newblock \emph{arXiv preprint arXiv:2501.12948}, 2025.

\bibitem[Hong et~al.(2024)Hong, Zhuge, Chen, Zheng, Cheng, Wang, Zhang, Wang, Yau, Lin, et~al.]{hongmetagpt}
Sirui Hong, Mingchen Zhuge, Jonathan Chen, Xiawu Zheng, Yuheng Cheng, Jinlin Wang, Ceyao Zhang, Zili Wang, Steven Ka~Shing Yau, Zijuan Lin, et~al.
\newblock Metagpt: Meta programming for a multi-agent collaborative framework.
\newblock In \emph{The Twelfth International Conference on Learning Representations}, 2024.

\bibitem[Huang et~al.(2025)Huang, Zhang, Wang, Qu, Lu, Roohani, Li, Qiu, Zhang, Di, et~al.]{biomni}
Kexin Huang, Serena Zhang, Hanchen Wang, Yuanhao Qu, Yingzhou Lu, Yusuf Roohani, Ryan Li, Lin Qiu, Junze Zhang, Yin Di, et~al.
\newblock Biomni: A general-purpose biomedical ai agent.
\newblock \emph{bioRxiv}, pages 2025--05, 2025.

\bibitem[Jin et~al.(2025{\natexlab{a}})Jin, Zeng, Yue, Yoon, Arik, Wang, Zamani, and Han]{search-r1}
Bowen Jin, Hansi Zeng, Zhenrui Yue, Jinsung Yoon, Sercan Arik, Dong Wang, Hamed Zamani, and Jiawei Han.
\newblock Search-r1: Training llms to reason and leverage search engines with reinforcement learning.
\newblock \emph{arXiv preprint arXiv:2503.09516}, 2025{\natexlab{a}}.

\bibitem[Jin et~al.(2025{\natexlab{b}})Jin, Zhang, Wang, and Cong]{stella}
Ruofan Jin, Zaixi Zhang, Mengdi Wang, and Le~Cong.
\newblock Stella: Self-evolving llm agent for biomedical research.
\newblock \emph{arXiv preprint arXiv:2507.02004}, 2025{\natexlab{b}}.

\bibitem[Li et~al.(2025{\natexlab{a}})Li, Dong, Jin, Zhang, Zhou, Zhu, Zhang, and Dou]{search-o1}
Xiaoxi Li, Guanting Dong, Jiajie Jin, Yuyao Zhang, Yujia Zhou, Yutao Zhu, Peitian Zhang, and Zhicheng Dou.
\newblock Search-o1: Agentic search-enhanced large reasoning models.
\newblock \emph{arXiv preprint arXiv:2501.05366}, 2025{\natexlab{a}}.

\bibitem[Li et~al.(2025{\natexlab{b}})Li, Jin, Dong, Qian, Zhu, Wu, Wen, and Dou]{webthinker}
Xiaoxi Li, Jiajie Jin, Guanting Dong, Hongjin Qian, Yutao Zhu, Yongkang Wu, Ji-Rong Wen, and Zhicheng Dou.
\newblock Webthinker: Empowering large reasoning models with deep research capability.
\newblock \emph{arXiv preprint arXiv:2504.21776}, 2025{\natexlab{b}}.

\bibitem[Li et~al.(2025{\natexlab{c}})Li, Zou, and Liu]{ToRL}
Xuefeng Li, Haoyang Zou, and Pengfei Liu.
\newblock Torl: Scaling tool-integrated rl.
\newblock \emph{arXiv preprint arXiv:2503.23383}, 2025{\natexlab{c}}.

\bibitem[Moonshot-AI(2025)]{kimi_researcher}
Moonshot-AI.
\newblock Kimi-researcher: End-to-end rl training for emerging agentic capabilities.
\newblock \url{https://moonshotai.github.io/Kimi-Researcher/}, 2025.
\newblock Accessed: 2025-06-26.

\bibitem[OpenAI(2022)]{chatgpt}
OpenAI.
\newblock Introducing chatgpt.
\newblock \url{https://openai.com/index/chatgpt/}, 2022.
\newblock Accessed: 2025-06-24.

\bibitem[{OpenAI}(2023)]{openai2023gpt4}
{OpenAI}.
\newblock Gpt-4 technical report.
\newblock \emph{arXiv preprint arXiv:2303.08774}, 2023.

\bibitem[OpenAI(2025{\natexlab{a}})]{o3}
OpenAI.
\newblock Introducing openai o3 and o4-mini.
\newblock \url{https://openai.com/index/introducing-o3-and-o4-mini/}, 2025{\natexlab{a}}.
\newblock Accessed: 2025-06-24.

\bibitem[OpenAI(2025{\natexlab{b}})]{o3-mini}
OpenAI.
\newblock Openai o3-mini.
\newblock \url{https://openai.com/index/openai-o3-mini/}, 2025{\natexlab{b}}.
\newblock Accessed: 2025-06-26.

\bibitem[OpenAI(2025{\natexlab{c}})]{openai_deep_research}
OpenAI.
\newblock Introducing deep research.
\newblock \url{https://openai.com/index/introducing-deep-research/}, 2025{\natexlab{c}}.
\newblock Accessed: 2025-06-26.

\bibitem[Ouyang et~al.(2022)Ouyang, Wu, Jiang, Almeida, Wainwright, Mishkin, Zhang, Agarwal, Slama, Ray, et~al.]{ouyang2022training}
Long Ouyang, Jeffrey Wu, Xu~Jiang, Diogo Almeida, Carroll Wainwright, Pamela Mishkin, Chong Zhang, Sandhini Agarwal, Katarina Slama, Alex Ray, et~al.
\newblock Training language models to follow instructions with human feedback.
\newblock \emph{NIPS}, 35:\penalty0 27730--27744, 2022.

\bibitem[Phan et~al.(2025)Phan, Gatti, Han, Li, Hu, Zhang, Zhang, Shaaban, Ling, Shi, et~al.]{hle}
Long Phan, Alice Gatti, Ziwen Han, Nathaniel Li, Josephina Hu, Hugh Zhang, Chen Bo~Calvin Zhang, Mohamed Shaaban, John Ling, Sean Shi, et~al.
\newblock Humanity's last exam.
\newblock \emph{arXiv preprint arXiv:2501.14249}, 2025.

\bibitem[Prabhudesai et~al.(2025)Prabhudesai, Chen, Ippoliti, Fragkiadaki, Liu, and Pathak]{prabhudesai2025maximizing}
Mihir Prabhudesai, Lili Chen, Alex Ippoliti, Katerina Fragkiadaki, Hao Liu, and Deepak Pathak.
\newblock Maximizing confidence alone improves reasoning.
\newblock \emph{arXiv preprint arXiv:2505.22660}, 2025.

\bibitem[Qian et~al.(2024)Qian, Liu, Liu, Chen, Dang, Li, Yang, Chen, Su, Cong, et~al.]{qian2024chatdev}
Chen Qian, Wei Liu, Hongzhang Liu, Nuo Chen, Yufan Dang, Jiahao Li, Cheng Yang, Weize Chen, Yusheng Su, Xin Cong, et~al.
\newblock Chatdev: Communicative agents for software development.
\newblock In \emph{Proceedings of the 62nd Annual Meeting of the Association for Computational Linguistics (Volume 1: Long Papers)}, pages 15174--15186, 2024.

\bibitem[Song et~al.(2025)Song, Jiang, Min, Chen, Chen, Zhao, Fang, and Wen]{r1-searcher}
Huatong Song, Jinhao Jiang, Yingqian Min, Jie Chen, Zhipeng Chen, Wayne~Xin Zhao, Lei Fang, and Ji-Rong Wen.
\newblock R1-searcher: Incentivizing the search capability in llms via reinforcement learning.
\newblock \emph{arXiv preprint arXiv:2503.05592}, 2025.

\bibitem[Sutton et~al.(1998)Sutton, Barto, et~al.]{sutton1998reinforcement}
Richard~S Sutton, Andrew~G Barto, et~al.
\newblock \emph{Reinforcement learning: An introduction}, volume~1.
\newblock MIT press Cambridge, 1998.

\bibitem[Tesauro et~al.(1995)]{tesauro1995temporal}
Gerald Tesauro et~al.
\newblock Temporal difference learning and td-gammon.
\newblock \emph{Communications of the ACM}, 38\penalty0 (3):\penalty0 58--68, 1995.

\bibitem[Wu et~al.(2025{\natexlab{a}})Wu, Li, Fang, Yin, Zhang, Tao, Zhang, Xi, Jiang, Xie, et~al.]{webdancer}
Jialong Wu, Baixuan Li, Runnan Fang, Wenbiao Yin, Liwen Zhang, Zhengwei Tao, Dingchu Zhang, Zekun Xi, Yong Jiang, Pengjun Xie, et~al.
\newblock Webdancer: Towards autonomous information seeking agency.
\newblock \emph{arXiv preprint arXiv:2505.22648}, 2025{\natexlab{a}}.

\bibitem[Wu et~al.(2025{\natexlab{b}})Wu, Zhu, and Liu]{agentic_reasoning}
Junde Wu, Jiayuan Zhu, and Yuyuan Liu.
\newblock Agentic reasoning: Reasoning llms with tools for the deep research.
\newblock \emph{arXiv preprint arXiv:2502.04644}, 2025{\natexlab{b}}.

\bibitem[Yang et~al.(2024)Yang, Yang, Zhang, Hui, Zheng, Yu, Li, Liu, Huang, Wei, Lin, Yang, Tu, Zhang, Yang, Yang, Zhou, Lin, Dang, Lu, Bao, Yang, Yu, Li, Xue, Zhang, Zhu, Men, Lin, Li, Xia, Ren, Ren, Fan, Su, Zhang, Wan, Liu, Cui, Zhang, and Qiu]{qwen2.5}
An~Yang, Baosong Yang, Beichen Zhang, Binyuan Hui, Bo~Zheng, Bowen Yu, Chengyuan Li, Dayiheng Liu, Fei Huang, Haoran Wei, Huan Lin, Jian Yang, Jianhong Tu, Jianwei Zhang, Jianxin Yang, Jiaxi Yang, Jingren Zhou, Junyang Lin, Kai Dang, Keming Lu, Keqin Bao, Kexin Yang, Le~Yu, Mei Li, Mingfeng Xue, Pei Zhang, Qin Zhu, Rui Men, Runji Lin, Tianhao Li, Tingyu Xia, Xingzhang Ren, Xuancheng Ren, Yang Fan, Yang Su, Yichang Zhang, Yu~Wan, Yuqiong Liu, Zeyu Cui, Zhenru Zhang, and Zihan Qiu.
\newblock Qwen2.5 technical report.
\newblock \emph{arXiv preprint arXiv:2412.15115}, 2024.

\bibitem[Ye et~al.(2025)Ye, Tang, Ge, Du, Yin, Chen, and Shao]{masgpt}
Rui Ye, Shuo Tang, Rui Ge, Yaxin Du, Zhenfei Yin, Siheng Chen, and Jing Shao.
\newblock Mas-gpt: Training llms to build llm-based multi-agent systems.
\newblock \emph{arXiv preprint arXiv:2503.03686}, 2025.

\bibitem[Zhang et~al.(2025)Zhang, Qiu, Wu, Li, Wang, Zhou, An, Chen, Li, Wang, et~al.]{trqa}
Zhongyue Zhang, Zijie Qiu, Yingcheng Wu, Shuya Li, Dingyan Wang, Zhuomin Zhou, Duo An, Yuhan Chen, Yu~Li, Yongbo Wang, et~al.
\newblock Origene: A self-evolving virtual disease biologist automating therapeutic target discovery.
\newblock \emph{bioRxiv}, pages 2025--06, 2025.

\bibitem[Zhao et~al.(2025{\natexlab{a}})Zhao, Wu, Yue, Wu, Xu, Lin, Wang, Wu, Zheng, and Huang]{zhao2025absolute}
Andrew Zhao, Yiran Wu, Yang Yue, Tong Wu, Quentin Xu, Matthieu Lin, Shenzhi Wang, Qingyun Wu, Zilong Zheng, and Gao Huang.
\newblock Absolute zero: Reinforced self-play reasoning with zero data.
\newblock \emph{arXiv preprint arXiv:2505.03335}, 2025{\natexlab{a}}.

\bibitem[Zhao et~al.(2025{\natexlab{b}})Zhao, Kang, Feng, Levine, and Song]{zhao2025learning}
Xuandong Zhao, Zhewei Kang, Aosong Feng, Sergey Levine, and Dawn Song.
\newblock Learning to reason without external rewards.
\newblock \emph{arXiv preprint arXiv:2505.19590}, 2025{\natexlab{b}}.

\end{thebibliography}
}

\newpage
\section{Cases}

\end{document}